# A Novel Tactile Force Probe for Tissue Stiffness Classification

Behafarid Darvish, Siamak Najarian, Elham Shirzad and Roozbeh Khodambashi
Artificial Tactile Sensing and Robotic Surgery Laboratory, Faculty of Biomedical Engineering,
Amirkabir University of Technology (Tehran Polytechnic), Tehran, Iran

**Abstract:** In this study, we have proposed a new type of tactile sensor that is capable of detecting the stiffness of soft objects. The sensor consists of a brass cylinder with an axial bore. An iron core can easily move inside the bore. Three peripheral bobbins were machined in the cylinder around which three coils have been wound. One of the coils was excited with an alternating current which caused a voltage to be induced in two other coils. A return spring was used to return the core to its initial position after it has been moved. The sensor was pressed against the surface of the object whose stiffness was going to be measured. The position of the core in this state was depended on the stiffness of the given object and the spring constant and was measured by measuring the change in the induced voltage in secondary coils. The proposed sensor was capable of measuring two contact parameters namely the applied force and the stiffness of the object. Using the data of this sensor, three different objects, made of polyurethane, silicon rubber and paraffin gel were discriminated. Thus, this sensor could be used in robot hands and minimally invasive surgery tools to improve their operation.

## INTRODUCTION

Tactile sensing is the detection and measurement of the spatial distribution of forces perpendicular to a predetermined sensory area and the subsequent interpretation of spatial information. It has the potential to have an impact on a large number of industries and disciplines. Key among these is the application to robotics in medicine and industrial automation[1]. Several types of tactile sensors have already been proposed for handling objects in robotics and automation systems. They can handle soft and fragile materials only with great difficulty[2]. In different biomedical engineering and medical robotics applications, tactile sensors can be used to sense a wide range of stimuli. This includes detecting the presence or absence of a grasped tissue/object or even mapping a complete tactile image[3-5].

Defining the state of manipulating or gripping of a biological tissue or an object requires the determination of two important physical parameters, i.e., force and position signatures[6-8]. Additionally, tactile and visual sensing is of great importance in different types of surgeries[9]. Minimally Invasive Surgery (MIS) is now being widely used as one of the most preferred choices for various types of operations[10-12]. In MIS, any inhibitions on the surgeon's sensory abilities might lead to undesirable results[13]. MIS has many advantages, including reducing trauma, alleviating pain, requiring smaller incisions, faster recovery time and reducing post-operation complications[14].

However, MIS decreases the tactile sensory perception of the surgeon. This effect is more pronounced during grasping or manipulation of biological tissues (i.e., veins, arteries, bones, etc.). In this regard, measuring the magnitude of the applied forces applied by the surgeon through the endoscopic graspers results in safer handling of biological tissues. The need to detect various tactile properties (such as stiffness, temperature and surface texture) justifies the key role of tactile sensing which is currently missing in MIS[9,15]. Present-day commercial endoscopic graspers do not have any built-in sensors, thus, the surgeon does not have the necessary tactile feedback to manipulate the tissue safely.

Stiffness is an important parameter in determining the physical properties of living tissue. Considerable biomedical attention has centered on the mechanical properties of living tissues at the single cell level. The Young's modulus of zona pellucida of bovine ovum was calculated using micro-tactile sensor fabricated and PZT material[16]. The stiffness of the cartilage of the

**Corresponding Author:** Siamak Najarian, Full-Professor of Biomedical Engineering, No. 424, Hafez Avenue, Department of Biomechanics, Faculty of Biomedical Engineering, Amirkabir University of Technology, Tehran, Iran
P.O. Box 15875-4413  Tel: (0098-21)-6454-2378  Fax: (0098-21)-6646-8186

human femoral condyles was measured via an ultrasonic tactile sensor under arthroscopic control[17]. The tactile sensor was useful for determining the intraoperative stiffness of healthy and diseased human cartilage in all grades. A new tactile sensor system has been developed for accurate measurement of myocardial stiffness in situ[18]. The design, fabrication, testing and mathematical modeling of a semiconductor microstrain gauge endoscopic tactile sensor have been investigated[13]. The sensor can measure, with reasonable accuracy, the magnitude and the position of an applied load on the grasper. Design and fabrication of piezoelectric-based tactile sensor for detecting compliance has been studied[19]. The sensor is capable of measuring the total applied force on the sensed object, as well as the compliance of the tissue/sensed object. Detecting the sensed objects compliance is based on the relative deformation of contact object/tissue on the rigid and compliant elements.

There has also been research on estimating the mechanical properties of the tissue through high-frequency shear deformations of the tissue sample and elastography techniques. A variety of other techniques also exist in the literature for estimating the viscoelastic characterization of tissues[20, 21]. Appearance of the stress contours of an embedded object in soft tissue on its surface as a result of difference between stiffness of tissue and the embedded object has been studied[22]. In this paper, we propose a new type of tactile sensor that can measure the applied force on its tip. When it is pressed against an object, the stress contours on the surface of the object change according to its stiffness and as a result, for different objects, different forces are measured by the sensor. Thus, the proposed sensor could be used to classify the objects using their stiffness. Flexibility and robustness along with the simplicity of design are the main advantages of this sensor over the previous ones.

## MATERIALS AND METHODS

The tactile sensor system consists of a tactile sensor, a data acquisition device and a personal computer for data analysis (Fig. 1). Since the output voltage of the sensor is small, it must be amplified before it can be sent to a computer for data processing. The data acquisition unit (Advantech PCI-1712) has been used for amplification and conversion of the analog signals to digital.

The sensor is pressed against the surface of the unknown object with the aid of a robot. According to the research performed by Hosseini et al.[22], this results in a stress distribution on object's surface which is proportional to the stiffness of the object. Therefore, the force that is measured with the sensor would be different for different objects.

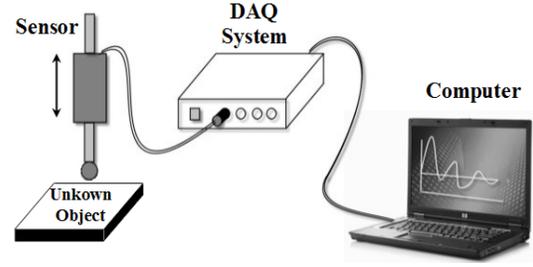

Fig. 1: Devices required for acquiring, processing and visualizing the output of the sensor

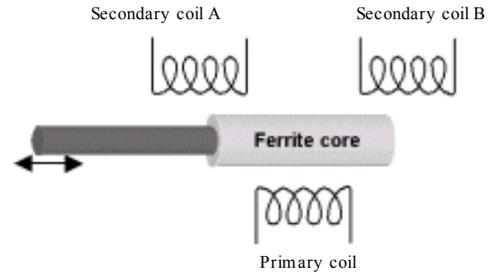

Fig. 2: A schematic of tactile sensor structure

**Sensor design**
**Sensor structure:** The tactile sensor operation is based upon the Faraday's induction law. According to this law, if a coil is placed in the varying magnetic field produced by another coil, a voltage will be induced in it which is proportional to the number of its turns and the rate of change of magnetic flux in it. This is given by:

$$\varepsilon = -N \frac{\partial \varphi}{\partial t} \qquad (1)$$

In the above equation, N is the number of turns of the coil, $\frac{\partial \varphi}{\partial t}$ is the rate of change of magnetic flux and ε is the induced voltage. The amount of flux passing the coil is related to the core material and the position of the core with respect to coil.

The tactile sensor is made up of one primary coil and two secondary coils which are shown by letters A and B in (Fig. 2). The primary coil is excited by a sinusoidal voltage that generates a varying magnetic field which causes a voltage to be induced in each secondary coil. These voltages are in phase but their magnitude is not equal and is calculated from Eq. 1. Since the numbers of turns of the secondary coils are equal, the induced voltage is solely proportional to the position of the core. At the center (where an equal length of core is inside each coil), the induced voltages are equal. When the core moves to the left, the voltage of the secondary coil A ($V_a$) will become greater than the voltage of the secondary coil B ($V_b$) and the difference of these voltages ($V_a$-$V_b$) is positive in sign. When the core moves to the right, the sign of the voltage difference becomes negative. Thus, the magnitude of potential difference indicates the amount of motion and its sign indicates the direction of motion.

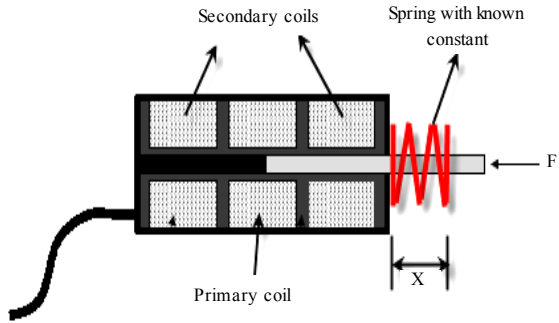

Fig. 3: Force measurement with the tactile sensor

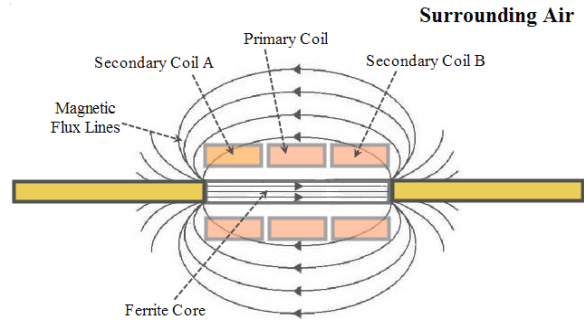

Fig. 4: Flux leakage in the sensor

For measuring the force exerted by the object when the sensor tip is brought into contact with it, a spring is added to the sensor as shown in Fig. 3. The displacement of the core due to exerted force is obtained from the change in output voltage of the secondary coils as discussed above and if the spring constant is known, the value of the exerted force could be calculated from:

$$F = k \times x \quad (2)$$

**Sensor modeling:** For a particular design, the geometry of the sensor such as core length, core diameter and number of turns of the primary and the secondary coil determines its performance which is characterized by several parameters like output function, sensitivity, resolution and nonlinearity. It is desired to know these parameters before manufacturing in order to obtain a design that best fits our requirements. Unfortunately, the electromagnetic relations are valid only for certain simple geometries. As shown in Fig. 4, there is a lot of flux leakage because the flux produced in the primary coil passes through the core and enters the surrounding air. Since the resistance to magnetic flux is high in the air, there is a considerable amount of decrease in the flux.

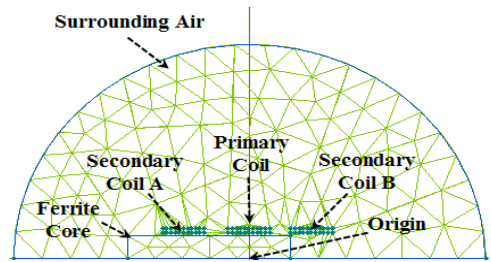

Fig. 5: Finite element model of the sensor

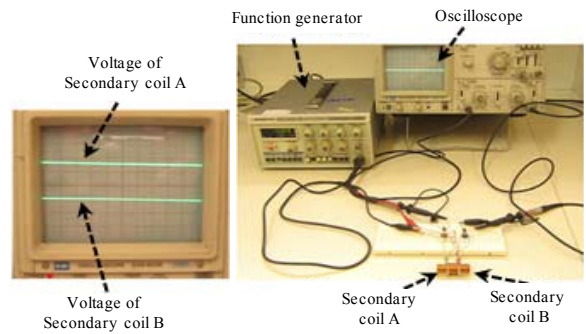

Fig. 6: Experiment setup

The best solution to account for the flux leakage is the use of finite element method. Finite element package Quickfield Ver. 5.5 has been used for modeling the sensor. This software is user-friendly and has a good accuracy. The geometry of the sensor is axisymmetric and thus only half of it has been modeled. Figure 5 shows the finite element model of the sensor.

The output voltage of the secondary coils when the core is at the origin is calculated. To validate the finite element results, this experiment is repeated with the manufactured sensor. The setup of this experiment is shown in Fig. 6. The primary coil is excited with a sinusoidal voltage that is generated with a function generator.

Table 1: Comparing finite element analysis and the experimental results

| Resistance (Ω) | Excitation voltage (V) | Output voltage (V) Experiment | Output voltage (V) FEM | Error (%) |
|---|---|---|---|---|
| 6.2 | 10 | 11 | 11.7 | 6.3 |
| 12.4 | 14 | 8 | 8.4 | 5 |
| 18.6 | 16 | 6 | 6.8 | 13 |

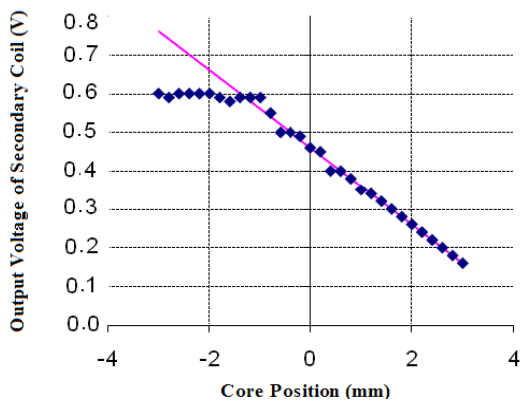

Fig. 7: Output voltage of the secondary coil A as a function of core position from the origin

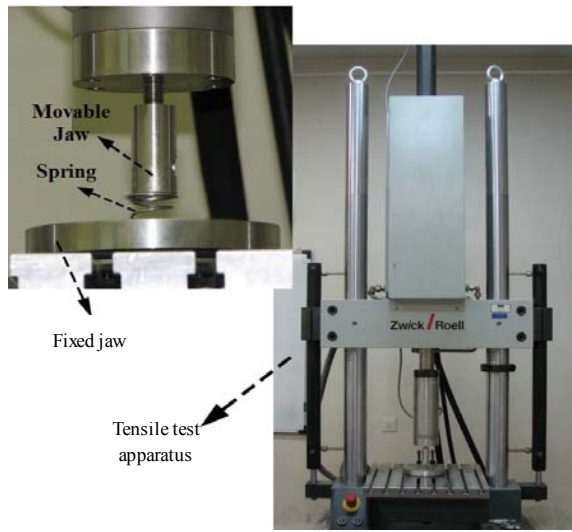

Fig. 8: Elongation test machine used to calibrate the spring

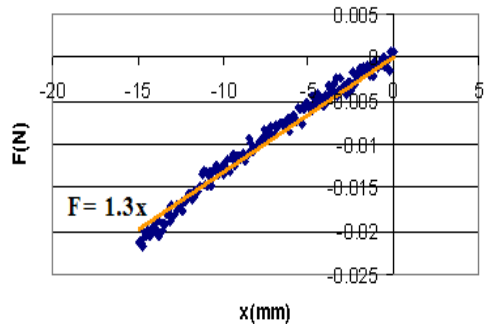

Fig. 9: Force-elongation curve of the spring

Table 1 compares the finite element analysis and the experimental results. The maximum error present is 13 percent.

To obtain the output function of the sensor, the core is first placed at the position shown in Fig. 5 i.e., it has passed all through the secondary coil A but is just about to enter the secondary coil B. The core is then moved to the right in steps of 0.2 mm and the finite element model has been solved for the output voltages of the secondary coils. The output voltage of the secondary coil A as a function of core position relative to origin is shown in Fig. 7. It can be seen that at the beginning, because the core is fully inserted in the secondary coil A, its voltage does not change with moving the coil. When the core reaches the edge of the coil and starts to leave it, its voltage starts to decrease. As seen, this decrease is linear. The slope of this curve is the sensitivity of the sensor and is calculated to be 0.1 mV/mm.

**Spring calibration:** In order to measure the applied force to the tip of the sensor, it is necessary to add a spring to it as shown in
Fig.Fig. 3. The spring constant should be determined with high accuracy. An elongation test machine (Zwick/Roell) was used to calibrate the spring (Fig. 8 Fig.). The spring was fixed between two jaws of the machine. The lower jaw is stationary but the upper jaw moves with a controllable rate. The force exerted by the spring is measured with the aid of force transducers in the machine. Fig. 9 shows the force-elongation curve of the spring. A straight line is fitted to the curve. The slope of this line is the desired spring constant which is calculated to be 1.3 N mm$^{-1}$.

## RESULTS AND DISCUSSION

To test the accuracy of the sensor in measuring the applied force, different forces were applied to it and the results of which are shown in Table 2. The maximum error present is 4.4%.

Three different materials were selected for classification test which are shown in Table 3. The specimens were of size 3×3 cm and a height of 2 cm. The sensor was indented on the surface of the specimens with a constant rate and was allowed to compress the surface for 1 mm. The force exerted by the specimens to the sensor tip and displacement of the tip were recorded by the sensor.

Table 2: Measured forces applied to the sensor

| Applied force (N) | Calculated force (N) | Error (%) |
|---|---|---|
| 2.501 | 2.599 | 3.9 |
| 2.845 | 2.972 | 4.4 |

Table 3: Materials selected for classification test

| Material | Young's modulus (MPa) |
|---|---|
| Paraffin gel | 0.77 |
| Silicon rubber | 1.07 |
| Polyurethane | 548 |

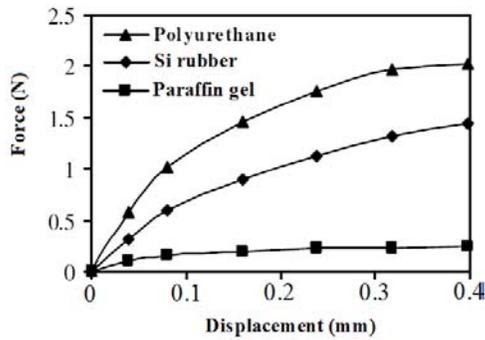

Fig. 10: Force-displacement curve for three different materials obtained by the sensor

Figure 10 shows the force-displacement curve of the materials given in the above table. It can be seen that for the same displacement, the force is higher in case of polyurethane compared to silicon rubber or paraffin gel and it can be concluded that its stiffness is higher.

## CONCLUSION

In this study, we have proposed a new type of tactile sensor that is capable of measuring the applied forces as it is in contact with other objects with high accuracy. The structure of the sensor is such that it can be mounted on endoscopic graspers. A major advantage of the designed system is that it can be easily miniaturized and micromachined.

The sensor is also capable of classifying the objects with respect to their stiffness. In this regard several materials were selected and tested and the sensor was able to classify them with high accuracy.

Using an array of the sensor can help determining the point of application of the force. Reducing the size of the sensor and using better signal processing algorithms are our future plans.

## ACKNOWLEDGEMENTS

We would like to express our gratitude to the Center of Excellence of Biomedical Engineering of Iran based in Amirkabir University of technology, Faculty of Biomedical Engineering for its contribution. We also thank Mr. A. Abouei Mehrizi for assisting us in the preparation of this manuscript.

## REFERENCES

1. Lee, M.H. and H.R. Nicholls, 1999. Tactile sensing for mechatronics-a state-of-the-art survey. Mechatronics, 9: 1-31. DOI: 10.1016/S0957-4158(98)00045-2.
2. Dargahi, J. and S. Najarian, 2005. Advances in tactile sensors design/manufacturing and its impact on robotics applications: A review. Ind. Robot, 32: 268-281. DOI: 10.1108/01439910510593965.
3. Dargahi, J., 2002. An endoscopic and robotic tooth-like compliance and roughness tactile sensor. J. Mech. Des., 124: 576-582, http://cat.inist.fr/?aModele=afficheN&cpsidt=13864081.
4. Najarian, S., J. Dargahi and X.Z. Zheng, 2006. A novel method in measuring the stiffness of sensed objects with applications for biomedical robotic systems. Int. J. Med. Robot. Comp., 2: 84-90. http://www.ncbi.nlm.nih.gov/pubmed/17520617.
5. Singh, H., J. Dargahi and R. Sedaghati, 2003. Experimental and finite element analysis of an endoscopic tooth-like tactile sensor. In: 2nd IEEE International Conference on Sensors, Oct. 22-24, Toronto, Canada, pp: 259-264. DOI: 10.1109/ICSENS.2003.1278939.
6. Dargahi, J., M. Parameswaran and S. Payandeh, 2000. A micromachined piezoelectric tactile sensor for an endoscopic grasper: Theory, fabrication and experiments. J. Microelectromech. Syst., 9: 329-335. DOI: 10.1109/84.870059.
7. Dargahi, J. and S. Payandeh, 1998. Surface texture measurement by combining signals from two sensing elements of a piezoelectric tactile sensor. In: Proceedings of the SPIE International Conference on Sensor Fusion: Architectures, Algorithms and Applications, April 16, Orlando, USA, pp: 122-128. DOI: 10.1117/12.303672.
8. Gray, B.L. and R.S. Fearing, 1996. A surface micro-machined micro-tactile sensor array. In:

Proceedings of IEEE International Conference on Robotic and Automation, April 22-28, Minneapolis, USA, pp: 1-6. DOI: 10.1109/ROBOT.1996.503564.

9. Howe, R.D., W.J. Peine, D.A. Kontarinis and J.S. Son, 1995. Remote palpation technology. IEEE Eng. Med. Biol. Mag., 14: 318-323. DOI: 10.1109/51.391770.
10. Dargahi, J. and S. Najarian, 2004. An integrated force-position tactile sensor for improving diagnostic and therapeutic endoscopic surgery. BioMed. Mater., 14: 151-166. http://iospress.metapress.com/content/0k8v09b0v9vpr6nw/.
11. Dargahi, J., M. Parameswaran and S. Payandeh, 2000. A micromachined piezoelectric tactile sensor for an endoscopic grasper-theory, fabrication and experiments. J. Microelectromech. Syst., 9: 329-335. Doi: 10.1109/84.870059.
12. Rao, N.P., J. Dargahi, M. Kahrizi and S. Prasad, 2003. Design and fabrication of a microtactile sensor. Canadian Conference on Electrical and Computer Engineering Towards a Caring and Human Technology. Montreal, Canada, 2: 1167-1170. http://ieeexplore.ieee.org/xpl/freeabs_all.jsp?arnumber=1226105.
13. Dargahi, J. and S. Najarian, 2003. An endoscopic force position grasper with minimum sensors. Can. J. Elect. Comput. Eng., 28: 151-161. DOI: 10.1109/CJECE.2003.1425102.
14. Dargahi, J.S. and Najarian, 2004. Analysis of a membrane type polymeric-based tactile sensor for biomedical and medical robotic applications. Sensor. Mater., 16: 25-41. http://sciencelinks.jp/j-east/article/200410/000020041004A0290021.php.
15. Dario, P., 1991. Tactile sensing-technology and applications, Sensor. Actuator. Phys., 26: 251-261. http://www-arts.sssup.it/people/prof/pdario/pdario.htm.
16. Murayama, Y., C.E. Constantinou and S. Omata, 2004. Micromechanical sensing platform for the characterization of the elastic properties of the ovum via uniaxial measurement. J. Biomech., 37: 67-72. doi:10.1016/S0021-9290(03)00242-2.
17. Uchio, Y., M. Ochi, N. Adachi, K. Kawasaki and J. Iwasa, 2002. Arthroscopic assessment of human cartilage stiffness of the femoral condyles and the patella with a new tactile sensor. Med. Eng. Phys., 24: 431-435. Doi: 10.1016/S1350-4533(02)00032-2.
18. Miyaji, K., S. Sugiura, H. Inaba, S. Takamoto and S. Omata, 2000. Myocardial tactile stiffness during acute reduction of coronary blood flow. Ann. Thorac. Surg., 69: 151-155. http://ats.ctsnetjournals.org/cgi/content/abstract/69/1/151.
19. Najarian, S., J. Dargahi, M. Molavi and H. Singh, 2006. Design and fabrication of piezoelectric-based tactile sensor for detecting compliance. In: IEEE International Symposium on Industrial Electronics, July 2006. 4: 3348-3352. DOI: 10.1109/ISIE.2006.296003.
20. A high-frequency shear device for testing soft biologicial tissues. J. Biomech., 30: 757-759. DOI: 10.1016/S0021-9290(97)00023-7.
21. Halperin, H.R. *et al*., 1991. Servo-controlled indenter for determining the transverse stiffness of ventricular muscle. IEEE T. Bio-Med. Eng., 38: 602-607. DOI: 10.1109/10.81586.
22. Hosseini, M., S. Najarian, S. Motaghinasab and J. Dargahi, 2006. Detection of tumours using a computational tactile sensing approach. Int. J. Med. Robot. Comp., 2: 333-340. http://www.ncbi.nlm.nih.gov/pubmed/17520652.